\documentclass{article}

\usepackage[preprint]{neurips_2025}

\usepackage[utf8]{inputenc} % allow utf-8 input
\usepackage[T1]{fontenc}    % use 8-bit T1 fonts
\usepackage{hyperref}       % hyperlinks
\usepackage{url}            % simple URL typesetting
\usepackage{booktabs}       % professional-quality tables
\usepackage{amsfonts}       % blackboard math symbols
\usepackage{nicefrac}       % compact symbols for 1/2, etc.
\usepackage{microtype}      % microtypography
\usepackage{xcolor}  
\usepackage[utf8]{inputenc} % allow utf-8 input
\usepackage[T1]{fontenc}    % use 8-bit T1 fonts
\usepackage{hyperref}       % hyperlinks
\usepackage{url}            % simple URL typesetting
\usepackage{booktabs}       % professional-quality tables
\usepackage{amsfonts}       % blackboard math symbols
\usepackage{nicefrac}       % compact symbols for 1/2, etc.
\usepackage{microtype}      % microtypography
\usepackage{xcolor}         % colors
\usepackage{hyperref}       % hyperlinks
\usepackage{url}            % simple URL typesetting
\usepackage{booktabs}       % professional-quality tables
\usepackage{amsfonts}       % blackboard math symbols
\usepackage{nicefrac}       % compact symbols for 1/2, etc.
\usepackage{microtype}      % microtypography
\usepackage{xcolor}         % colors
\usepackage{amsmath}
\usepackage{amssymb}
\usepackage{amsbsy}
\usepackage{dsfont}
\usepackage{multirow}
\usepackage{amsthm}
\usepackage{makecell}
\usepackage{pifont}
\usepackage{graphicx}
\usepackage{svg}
\usepackage{algorithm} 
\usepackage{algorithmic}
\usepackage{xparse}% colors

\title{Auditing Algorithmic Bias in Transformer-Based Trading}

\author{
  Armin Gerami, Ramani Duraiswami\\
  Department of Computer Science, Umiacs\\
  University of Maryland\\
  College Park, MD \\
  \texttt{[agerami, ramanid]@umd.edu}
}

\begin{document}

\maketitle

\vspace{-10pt}
\begin{abstract}
Transformer models have become increasingly popular in financial applications, yet their potential risk making and biases remain under-explored. The purpose of this work is to audit the reliance of the model on volatile data for decision-making, and quantify how the frequency of price movements affects the model's prediction confidence. We employ a transformer model for prediction, and introduce a metric based on Partial Information Decomposition (PID) to measure the influence of each asset on the model's decision making. Our analysis reveals two key observations: first, the model disregards data volatility entirely, and second, it is biased toward data with lower-frequency price movements.
\end{abstract}

\section{Introduction}
\vspace{-3pt}
The recent advancements of advanced artificial intelligence is instigating a paradigm shift in quantitative finance. The auto-regressive Transformer models~\cite{vaswani2017attention}, which have powered recent breakthroughs in natural language processing and computer vision, are now being rapidly adopted for algorithmic stock trading~\cite{coelho2024transformers}. Architectures such as the "Quantformer"~\cite{zhang2024quantformer} leverage sophisticated self-attention mechanisms to capture complex, long-range dependencies within noisy financial time-series, demonstrating empirical superiority over predecessor models like Long Short-Term Memory (LSTM) networks. These models achieve higher predictive accuracy and generate superior cumulative returns, establishing a new state-of-the-art in financial forecasting~\cite{guo2025trading}. However, their complexity gives rise to a critical challenge: the models operate as opaque "black boxes," where the intricate path from data input to trading decision is largely an enigma, even to their developers~\cite{umeaduma2025explainable}. This lack of transparency poses significant risks for accountability, fairness, and regulatory compliance, creating a need for rigorous auditing methodologies.

\par This paper trains a Transformer model to predict stock price movements and investigates two core aspects of its decision-making process. First, we audit the trustworthiness of the model's decisions by measuring whether a stock's implied volatility (IV) affects the model's reliance on that stock. Second, we perform an ablation study in which we control the frequency of price movements to simulate various trading speeds for each stock and then measure how this frequency affects the model's reliance on it.

\section{Background}
\vspace{-3pt}
This section outlines the analytical frameworks used in our study; Transformer architecture, and PID.

\subsection{Transformer Models}
\vspace{-3pt}
Transformers were designed to overcome the critical limitations of prior recurrent architectures, such as Recurrent Neural Networks (RNNs) and Long Short-Term Memory (LSTM) networks. Specifically, the sequential nature of RNNs precludes parallelization within training examples and makes it challenging to capture long-range dependencies due to vanishing gradients. The Transformer architecture solves these issues by relying on a parallelizable self-attention mechanism.

\par The core innovation of the Transformer is the attention mechanism, which allows the model to weigh the importance of all other tokens in an input sequence when computing a representation for a given token. This mechanism operates on a set of queries ($Q)$, keys ($K$), and values ($V$), which are linear projections of the input embeddings. To enhance the model's representational power, the Transformer employs Multi-Head Attention, which runs the self-attention mechanism multiple times in parallel with different, learned linear projections for the $Q$, $K$, and $V$ vectors. This allows each "head" to focus on different types of relationships within the sequence. 

\par Given a sequence of $N$ tokens, model dimension of $C$, $H$ heads, and dimension per head of $D = C/H$, the output of each attention head $O$ is derived as 
\begin{gather}
  \mathbf{O} = \mathbf{A}\mathbf{V},\,\, \mathbf{A} = \mbox{Softmax}\left(\mathbf{Q}\mathbf{K}^T\right),\\
        \mathbf{o_{ij}} \!=\! \dfrac{\sum_{n=1}^{N}{\exp(\mathbf{q}_i.\mathbf{k_n}/\sqrt{D})\mathbf{v_{n,j}}}}{\sum_{n=1}^{N}{\exp(\mathbf{q}_i.\mathbf{k}_n/\sqrt{D})}} \!=\! \dfrac{\sum_{n=1}^{N}{f(\mathbf{q}_i.\mathbf{k_n})\mathbf{v_{n,j}}}}{\sum_{n=1}^{N}{f(\mathbf{q}_i.\mathbf{k}_n)}}\label{0}
\end{gather}

Then, the output of all the attention heads are concatenated and linearly projected to form the final output of the attention layer.

\subsection{Partial Information Decomposition}
\vspace{-3pt}
\label{pid}
PID~\cite{kolchinsky2022novel} is a theoretical framework designed to analyze how a set of input variables provides information about an output variable. Its goal is to move beyond simple mutual information and dissect how the information is structured among the sources. The central idea is to break down the total information into distinct, non-overlapping components, each describing a different mode of interaction:
\begin{itemize}
    \item \textbf{Mutual Information ($I$)} The mutual information between a source and the target variable.
    \item \textbf{Union Information ($U$):} The mutual information provided by at least one individual source.
    \item \textbf{Excluded Information ($E$):}  The information in the union of the sources except one particular source.
\end{itemize}

\par Given source variables $X_1, X_2, ..., X_k$ and target variable $Y$, the PID is written as
\begin{gather}
    E(X_i\rightarrow Y\,|\,X_1, X_2, ..., X_k) = U(X_1, X_2, ..., X_k;\,Y) - I(X_i;\, Y),\\
    I(X_i;\, Y) = H(X_i) - H(X_i|Y),\quad H(.) := \text{Shannon Entropy}\\
    U(X_1, X_2, ..., X_k;\,Y) = \underset{Q}{\text{Inf }} I(Q;\,Y)\,\text{such that}\, \forall i\, X_i \in Q,
\end{gather}
where $E(X_i\rightarrow Y\,|\,X_1, X_2, ..., X_k)$ measures "What knowledge does the rest of the group have that $X_i$ is missing".

\section{Method}
\vspace{-3pt}
First, we explain how we train our transformer to predict the price movement of a 'target' stock. The model uses the price time-series of the target stock and other relevant 'support' stocks to gain a broader perspective on market movements. Then, we describe how we use PID to measure the influence of each support stock on the transformer's predictions for the target stock.

\subsection{Training Transformer}
\vspace{-3pt}
We use historical data from \textbf{January 2025} to \textbf{June 2025} for training and validation, using the percentage return ($pr$) of each stock as the input data. The $pr$ is calculated as:
\begin{gather}
    pr(t) = \dfrac{price(t) - price(t-1)}{price(t)}.
\end{gather}
For simplicity, we only consider $pr$, although utilizing additional indicators might improve accuracy. We use \textbf{NVDA} as the target stock and \textbf{AMD}, \textbf{MU}, \textbf{TSM}, and \textbf{INTC} as the support stocks. The $pr$ and IV of these stocks can be found in Appendix~\ref{app:pr}. 

\par Thus, the input to the transformer at each timestep is a five-dimensional vector containing the $pr$ of these five stocks. This vector is then passed to the first attention layer, and we set the timestep as 1 hour. The output of each attention layer is then forwarded to the next, with the final layer producing a 64-dimensional vector. This vector represents a probability distribution over quantized predicted $pr$ values in the range of $[-12.8\%, 12.8\%]$, similar to how an LLM's output is a probability distribution over the possible tokens. This design was chosen over directly predicting a single $pr$ value because a probability distribution is necessary for calculating mutual information during PID. Figure~\ref{fig:prediction} in Appendix~\ref{app:pr} shows the model's prediction result. We should emphasize that focus of this study is on the transformer's decision-making process, and not it's predictive accuracy.

\par The network has an embedding size of $D=256$, $H=8$ attention-heads per attention layer, and four attention layers. Furthermore, we use ROPE~\cite{pochet2023roformer} for positional encoding, and take advantage of add-and-norm residual connection. We set the context length to $N=64$, meaning the model uses the $pr$ of the target and support stocks from the previous $64$ timesteps to predict the target's $pr$ for the next timestep. We use mean Cross-Entropy as our loss function, and apply a dropout rate of $0.1$ to prevent overfitting.

\subsection{Measuring Influence}
\vspace{-3pt}
Our goal is to: 1. Measure the influence of each support stock on our transformer model's prediction of the target stock's $pr$. 2. Determine whether the model has less reliance on support stocks with higher IV. 3. Investigate a potential bias in the model towards the trading frequency of the stocks. Let's denote the target and support stocks as $X_1,..., X_5$, and the model's prediction as $Y$.

\par We measure the influence of $X_i$ by calculating the excluded information, $E(X_i\rightarrow Y|X)$, using PID. The lower the excluded information, the higher the model's dependence on $X_i$, and therefore the higher its influence. To derive excluded information, a distribution for the stocks' $pr$ and the model output is needed. We approximate the distribution of the input through quantizing the $pr$s and deriving a histogram over the six month time frame. The probability distribution of the model output is given by the transformer itself, as the final attention layer generates a probability for each value within a predefined set of possible $pr$s.

% \par The IV of each stock can be derived using the Black-Scholes model, and is available as historical market data\footnote{We used Yahoo Finance Python library.}. Ideally, our model should have lower reliance on stocks with a higher IV. Therefore, the supports with higher IV should have higher excluded information. However, as we will explain in Section~\ref{result}, this is not the case. To investigate this unwanted bias, we then examine how trading frequency affects this behavior. Since higher trading frequency can cause greater price variation, we apply a low-pass filter to the price time-series to synthetically smooth these fluctuations, simulating a lower frequency of trades. We find in Section~\ref{result} that applying this filter to a support stock causes its excluded information to increase, which indicates an undesirable model bias towards stocks with higher trading frequencies.

\vspace{-3pt}
\section{Results}
\vspace{-3pt}
\label{result}
This section presents two key observations from our results: 1. High IV does not discourage the transformer model from relying on a support stock. 2. The model exhibits a bias toward data from stocks with higher trading frequencies. Ideally, our model should have lower reliance on stocks with a higher IV. Therefore, the supports with higher IV should have higher excluded information. However, as we will show in Section~\ref{4.1}, this is not the case. We then examine how trading frequency affects the model's decision making. Since higher trading frequency can cause greater price variation, we apply a low-pass filter to the price time-series to synthetically smooth these fluctuations, simulating a lower frequency of trades. We find in Section~\ref{4.2} that applying this filter to a support stock causes its excluded information to decrease, which indicates an unwanted model bias towards stocks with lower trading frequencies.

\subsection{Reliance Independent of IV}
\label{4.1}
\begin{figure*}[!h]
\minipage{0.499\linewidth} \begin{centering}
  \includegraphics[width=\linewidth, trim=0 10pt 0 5pt, clip=true]{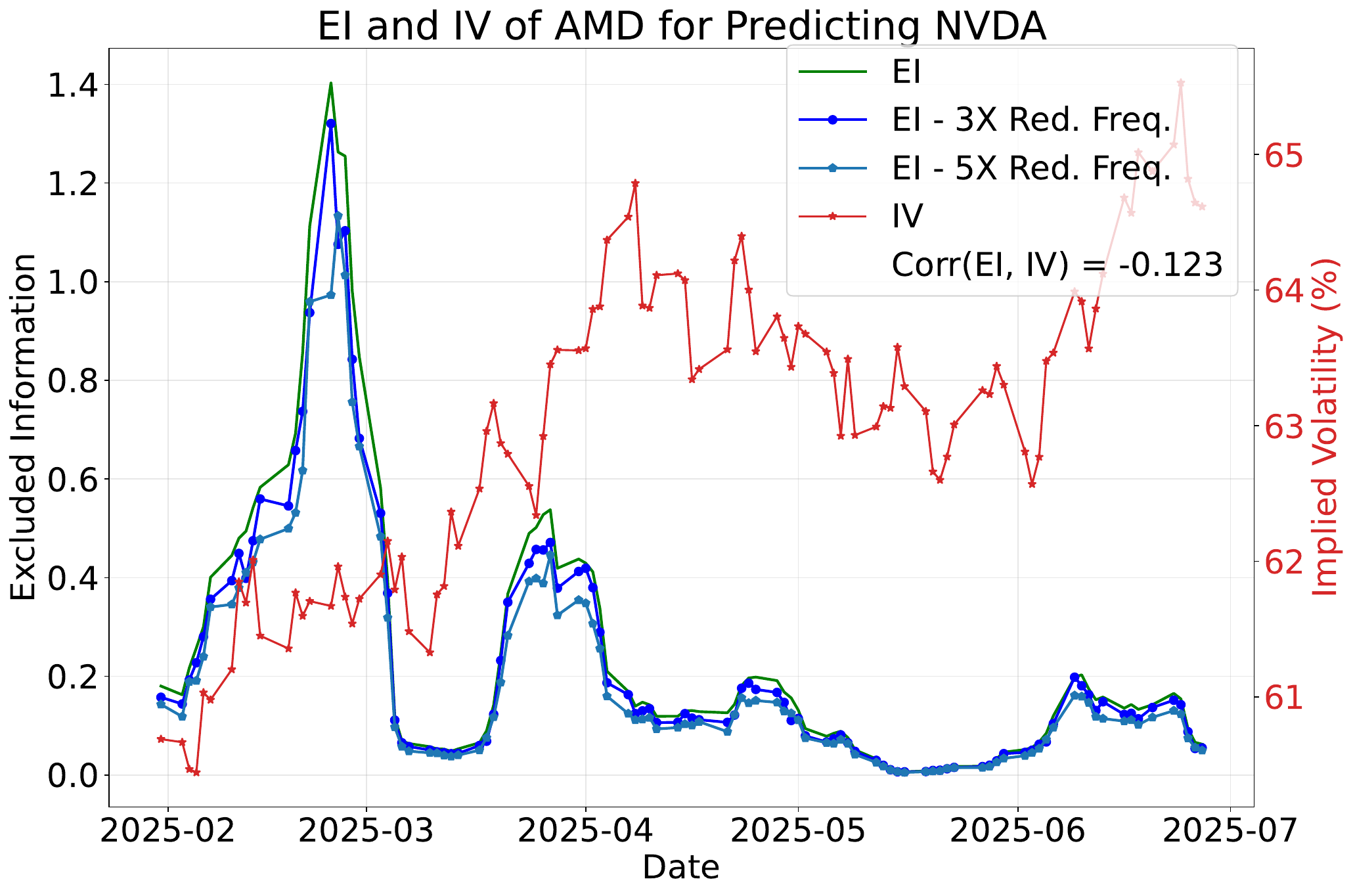}
  \end{centering} \endminipage
\minipage{0.499\linewidth} \begin{centering}
  \includegraphics[width=\linewidth, trim=0 10pt 0 5pt, clip=true]{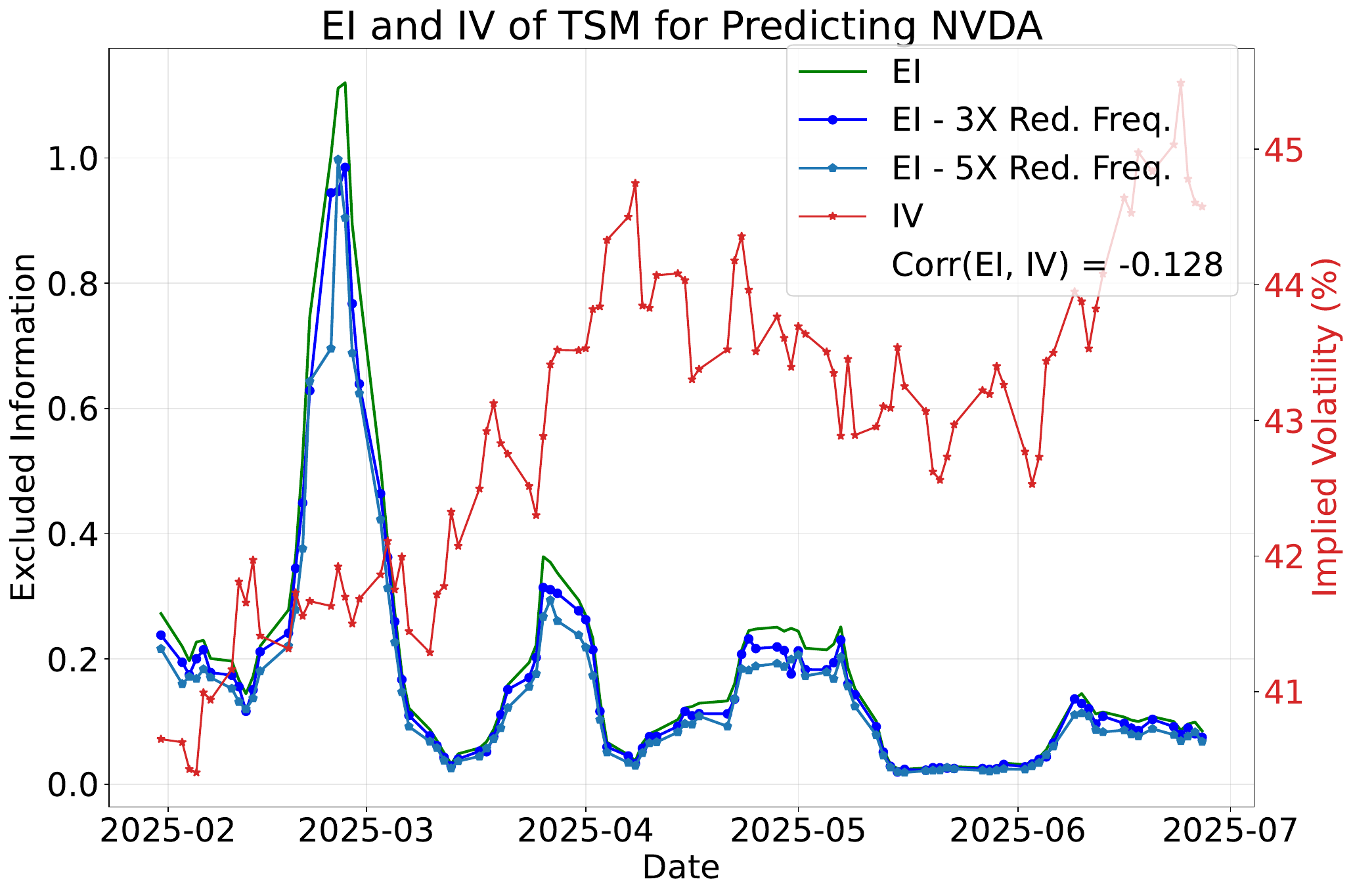}
  \end{centering} \endminipage
\vspace{0pt}
\minipage{0.499\linewidth} \begin{centering}
  \includegraphics[width=\linewidth, trim=0 10pt 0 5pt, clip=true]{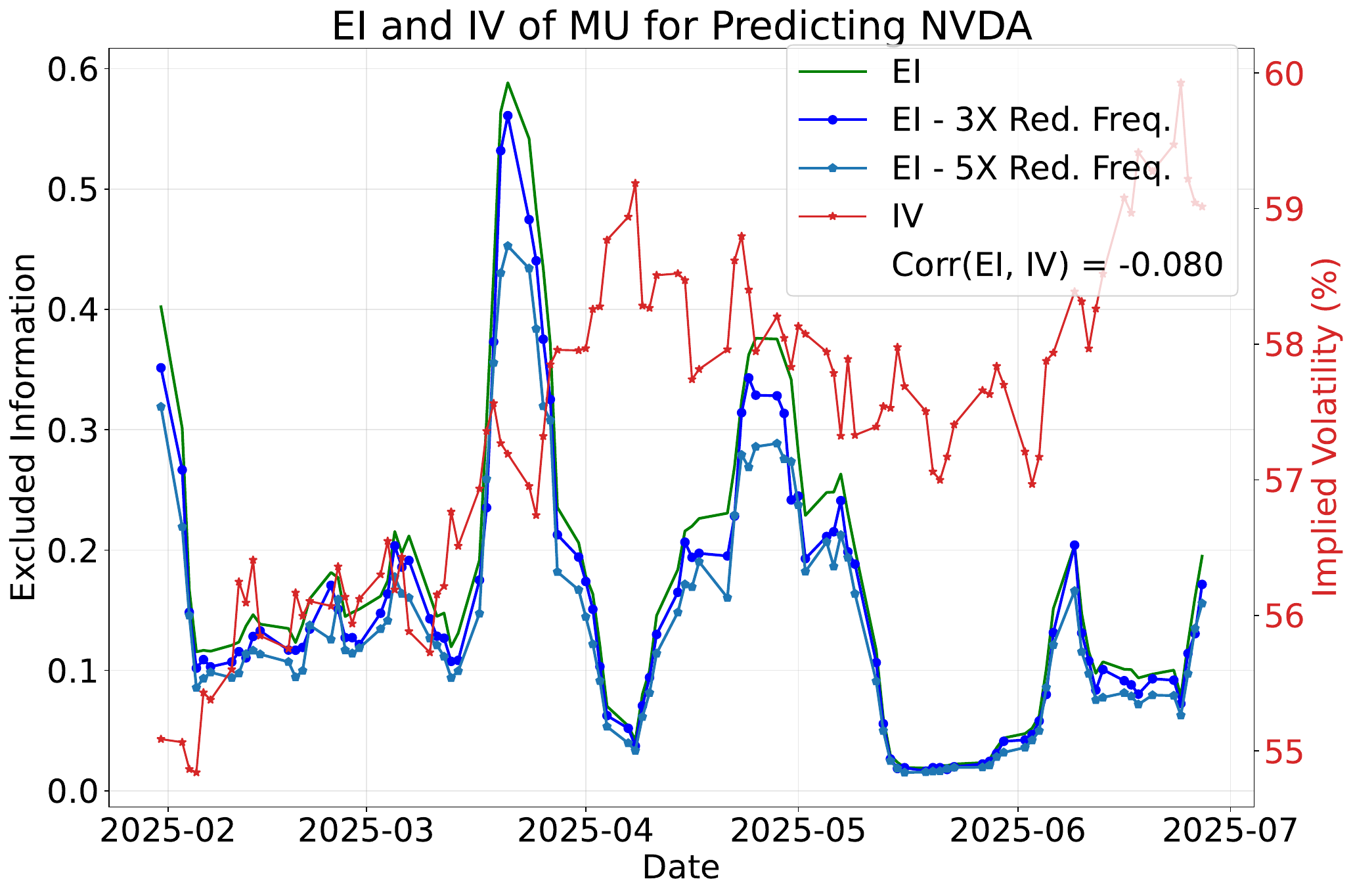}
  \end{centering} \endminipage
\minipage{0.499\linewidth} \begin{centering}
  \includegraphics[width=\linewidth, trim=0 10pt 0 5pt, clip=true]{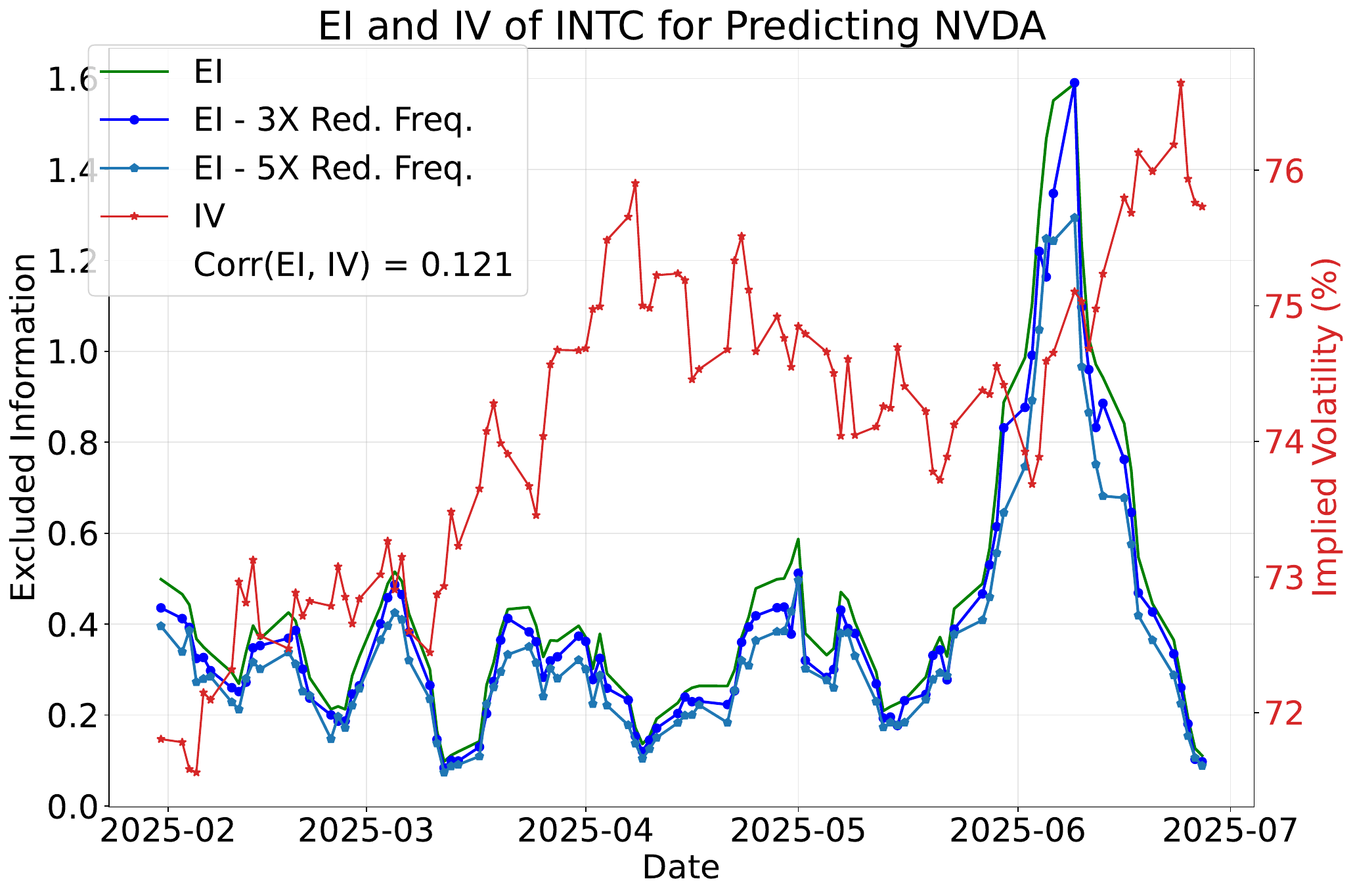}
  \end{centering} \endminipage
  \vspace{-20pt}
\caption{Excluded Information (EI) for each support stock used when predicting the target stock, NVDA. A higher EI value indicates that the model relies less on that stock's data. The right axis displays the 30-day implied volatility (IV) as a proxy for risk. Ideally, high IV should lead to high EI (less reliance on risky data), but the near-zero correlation shown here indicates the model is not discounting this risk. The plot also includes scenarios where support stocks are modified to simulate 3× and 5× reduced trading frequencies for the support stocks, which consistently results in lower EI.}
\vspace{-3pt}
\label{fig:result}
\end{figure*}

\vspace{-3pt}
In Figure~\ref{fig:result} we plot the IV for each of the support stocks, and their excluded information on daily average. Excluded information measures the model's reliance on that stock for the target prediction, and a higher value means lower reliance. We want the model to rely less on stocks with higher IV. However, the results show that the transformer's reliance on a stock is independent of its IV. There are numerous instances that the IV is high while EI is high as well, and the overall correlation between Ei and IV is close to zero. Ideally the EI should be high whenever IV is high to indicate that the model has reduced reliance on a data when the uncertainty is high, and the correlation should be close to $1$.

\par Relying on a support stock with high IV is risky since its price is more likely to experience large, sudden swings~\cite{addy2024algorithmic}. These movements can influence the model, causing its prediction to mirror the support stock's trajectory. If this movement is specific to that stock and not reflective of the broader market, it can lead the model to predict an incorrect $pr$ direction. For example, the model relies most heavily on \textbf{AMD} (as shown by its low excluded information) despite it having the highest IV. If company-specific news were to cause a sharp drop in \textbf{AMD}'s price, this high reliance would skew the model's prediction for \textbf{NVDA} downward, even if the news had no bearing on \textbf{NVDA}'s actual value.

\subsection{Trade Frequency Bias}
\label{4.2}
\vspace{-3pt}
To evaluate the effect of trade frequency of the support stocks on our model, we train the model under three sets of condition: 1. Use the same price movement timestep as the target stock (1 hour timesteps), 2. use a $3\times$ reduced frequency (3 hour timesteps), and 3. use $5\times$ reduced frequency (5 hour timesteps). Figure~\ref{fig:result} shows the daily average of EI of the support stocks for each of these conditions, as well as the IV. The results show that the $5\times$ reduced frequency consistently achieves lower EI, followed by the $3\times$ reduced frequency.
\par This indicates that the model prioritizes lower fluctuations, which can be attributed to two phenomena. First, reducing the price movement frequency essentially applies a low-pass filter, which may result in the removal of high-frequency noise. Second, this filtering can also mitigate overfitting. This is because the model avoids learning random patterns as if they were a genuine pattern, which would cause it to perform poorly on new data.
%%%%%%%%%%%%%%%%%%%%%%%%%%%%%%%%%%%%%%%%%%%%%%%%%%%%%%%%%%%%

\section{Conclusion}
\vspace{-3pt}
In conclusion, this paper takes a step to address the critical "black box" problem of Transformer models in finance, moving towards understanding the logic of these powerful but opaque models. We investigated the model's decision-making by auditing its response to risk (measured via implied volatility) and analyzing its sensitivity to the frequency of price movements through an ablation study. We found that the model disregards data volatility and is biased toward data with lower-frequency price movements.
%%%%%%%%%%%%%%%%%%%%%%%%%%%%%%%%%%%%%%%%%%%%%%%%%%%%%%%%%%%%

\newpage
\bibliography{main}

\newpage
\appendix
\section{Supporting Figures}
\begin{figure*}[!h]
\hspace{95pt}\minipage{0.499\linewidth} \begin{centering}
  \includegraphics[width=\linewidth, trim=0 0 0 0, clip=true]{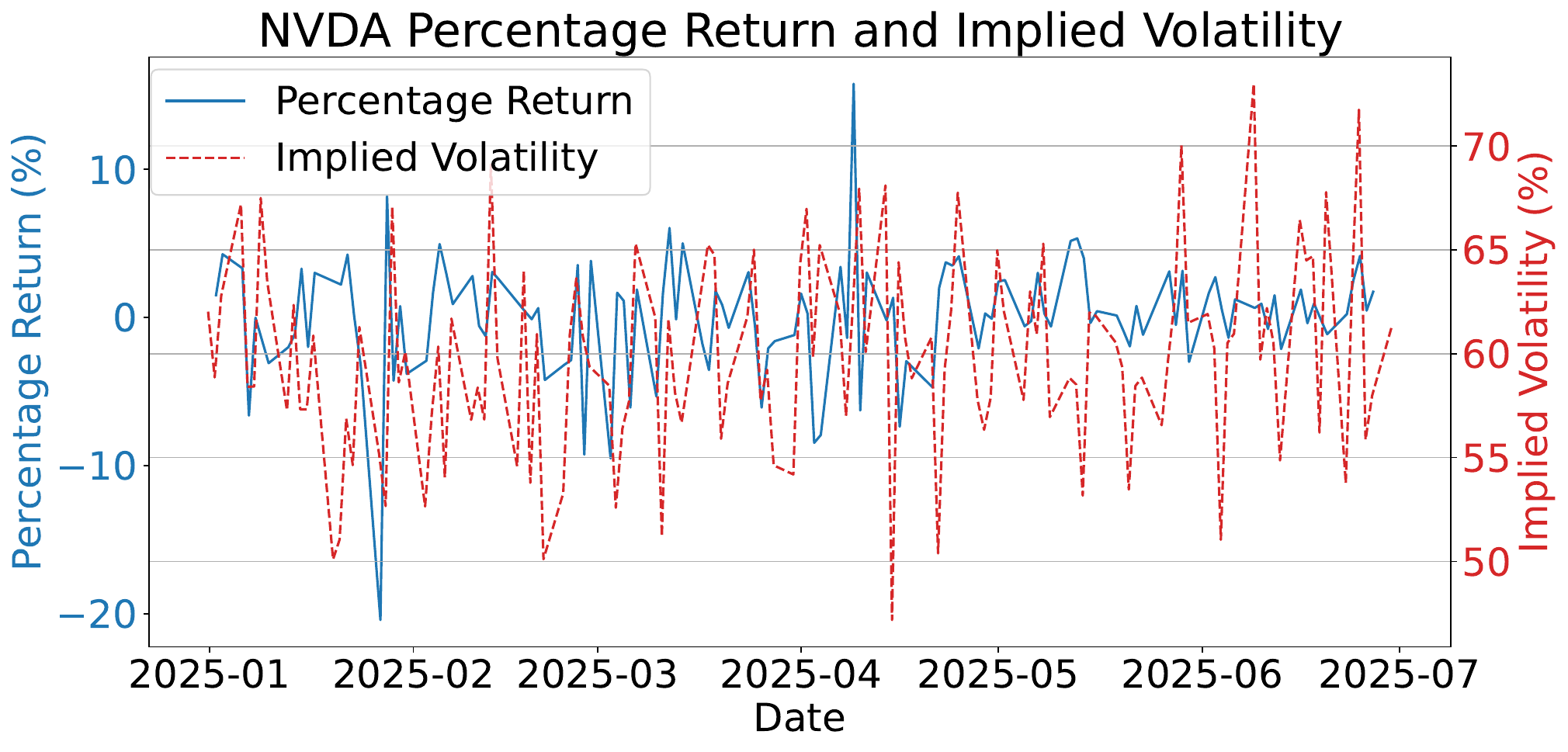}
  \end{centering} \endminipage\hfill
\vspace{1pt}
\minipage{0.499\linewidth} \begin{centering}
  \includegraphics[width=\linewidth, trim=0 0 0 0, clip=true]{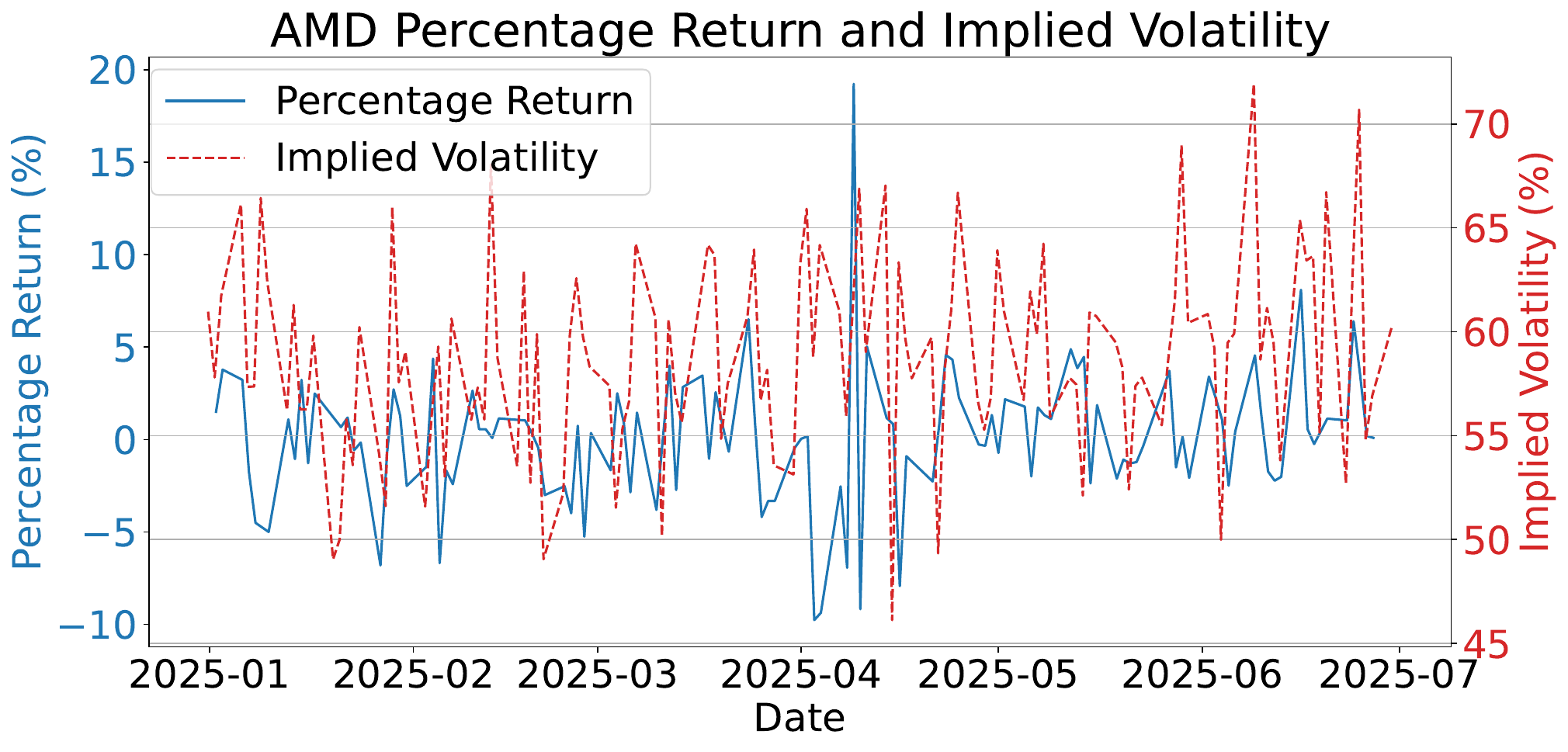}
  \end{centering} \endminipage
\minipage{0.499\linewidth} \begin{centering}
  \includegraphics[width=\linewidth, trim=0 0 0 0, clip=true]{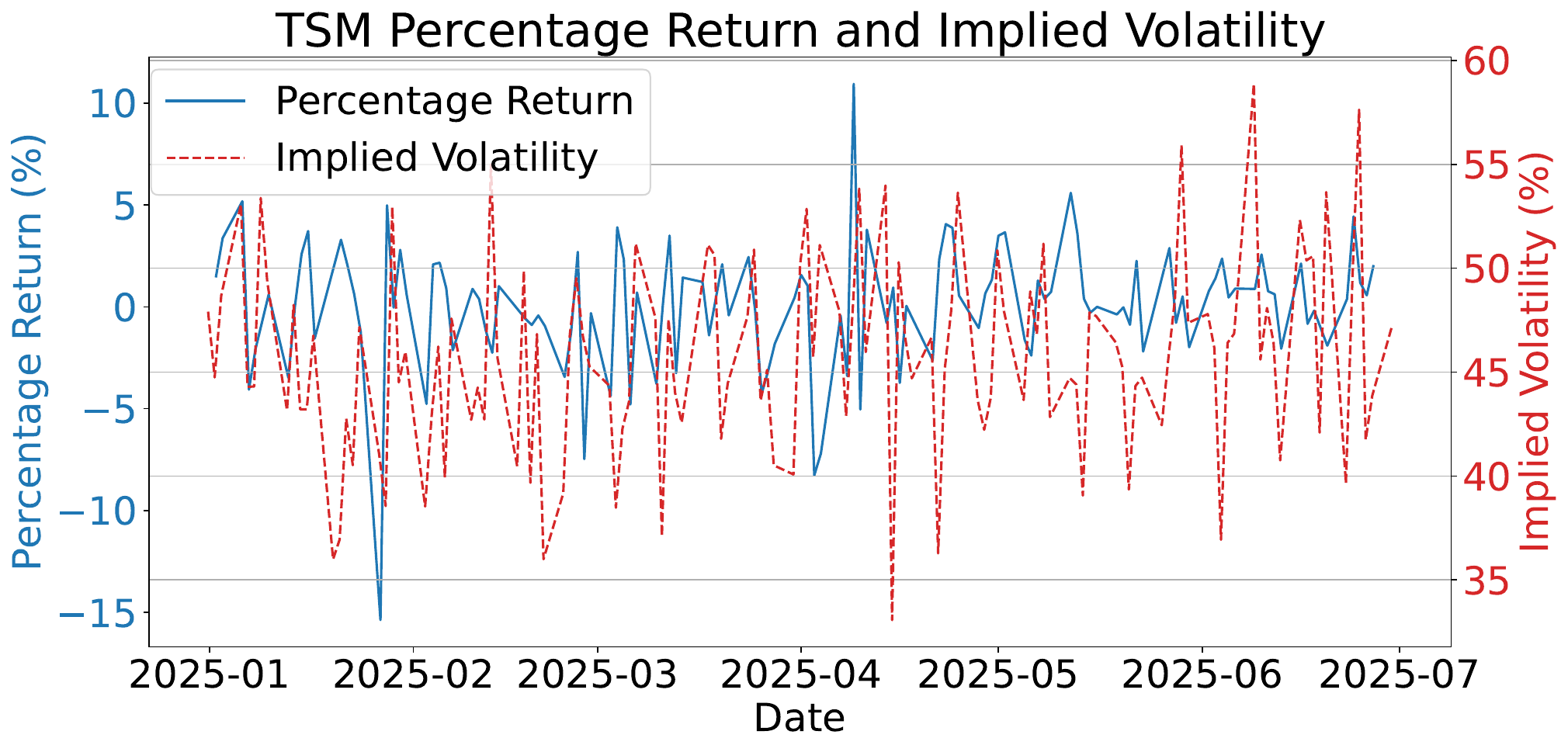}
  \end{centering} \endminipage
\vspace{0pt}
\minipage{0.499\linewidth} \begin{centering}
  \includegraphics[width=\linewidth, trim=0 0 0 0, clip=true]{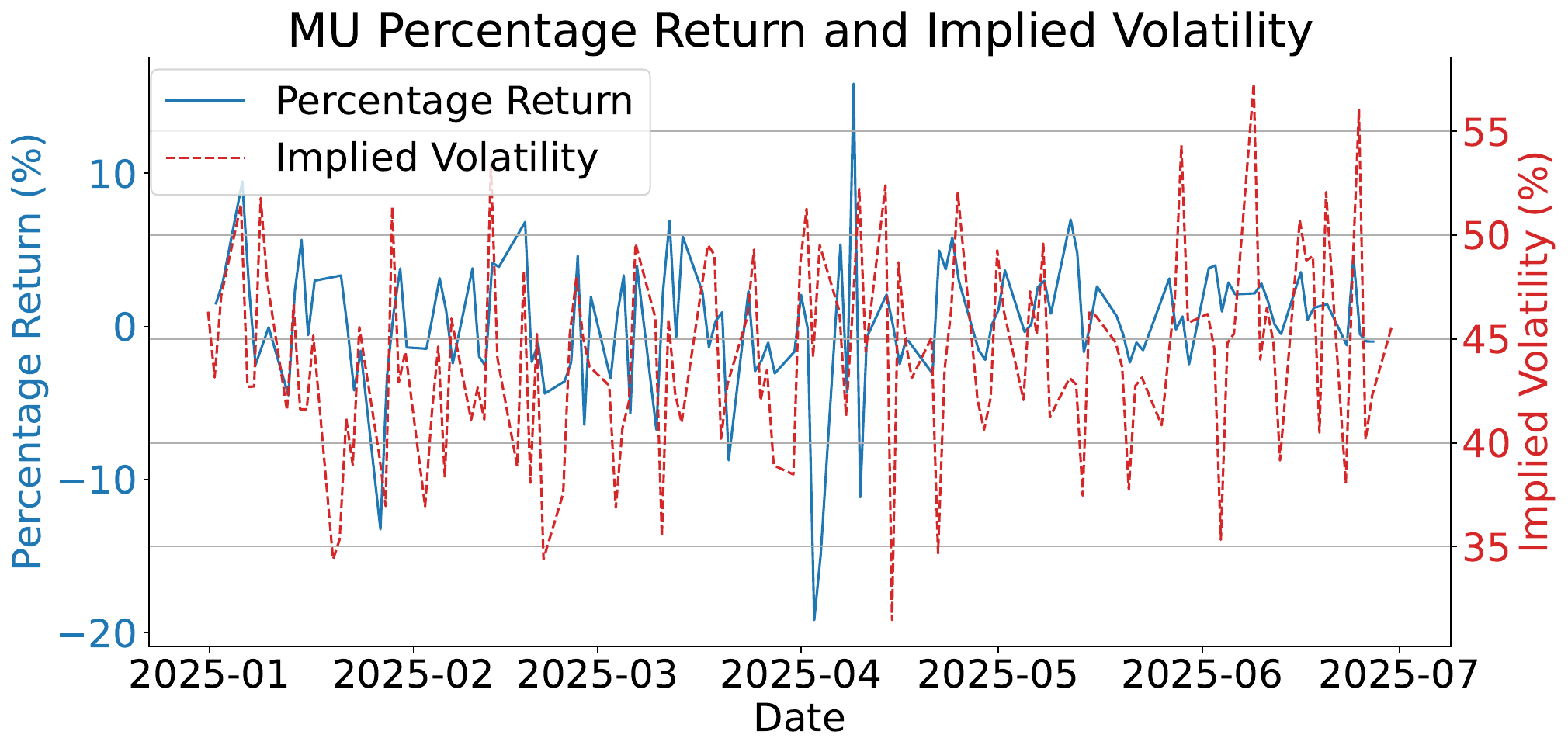}
  \end{centering} \endminipage
\minipage{0.499\linewidth} \begin{centering}
  \includegraphics[width=\linewidth, trim=0 0 0 0, clip=true]{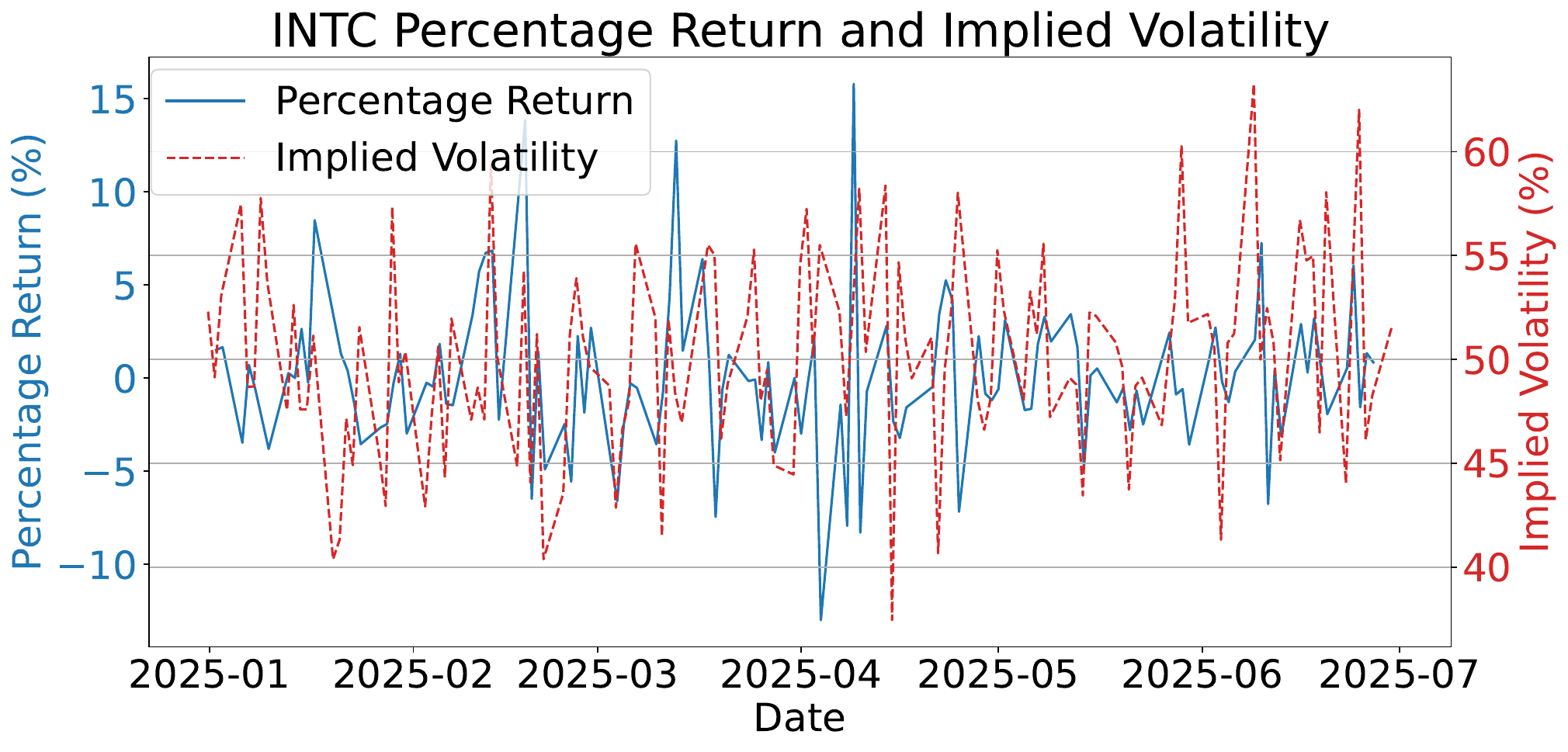}
  \end{centering} \endminipage
\caption{Percentage return and 30-day implied volatility of our target stock (NVDA), and the support stocks (AMD, TSM, MU, INTC).}
\label{fig:stocks}
\end{figure*}

The appendix includes figures to illustrate percentage return ($pr$) and implied volatility (IV) for the target stock NVDA and each of the support stocks in Figure~\ref{fig:stocks}.
\par
\begin{figure*}[!h]
\begin{centering}
  \includegraphics[width=0.7\linewidth, trim=0 0 0 0, clip=true]{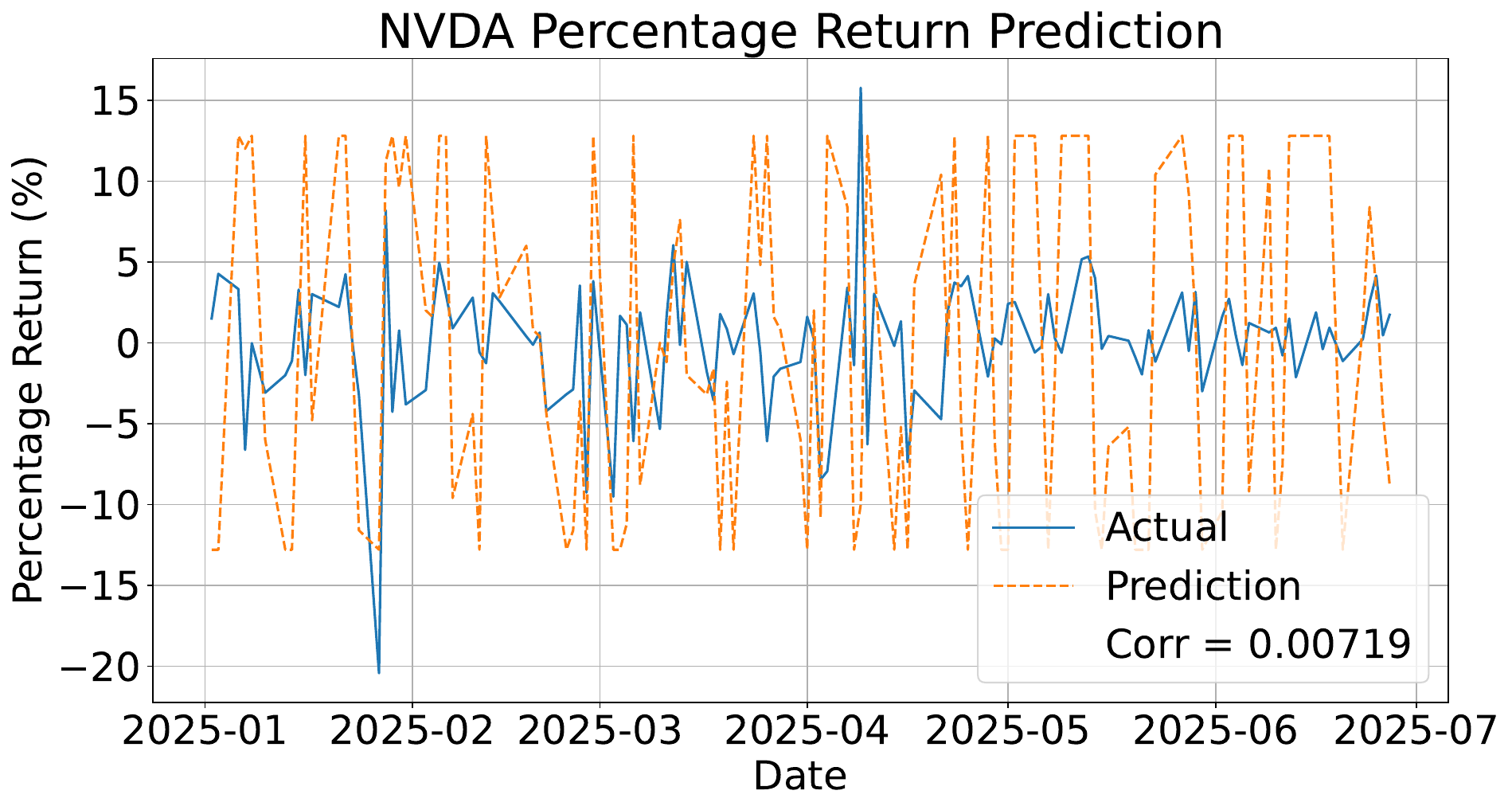}
\caption{The actual percentage return of the NVDA stock and the trained transformer's prediction.}
\label{fig:prediction}
\end{centering}
\end{figure*}

 Furthermore, Figure~\ref{fig:prediction} shows the trained Transformer model's $pr$ prediction and actual data. The high correlation indicates meaningful model prediction.
\label{app:pr}

\end{document}